\documentclass[11pt]{article}
\usepackage{authblk}

\newlength{\defbaselineskip}
\usepackage{multirow}
\usepackage{booktabs}
\usepackage{url}
\usepackage{graphicx}
\usepackage{amsmath}
\usepackage{amssymb}
\usepackage[caption=false,font=footnotesize]{subfig}

\usepackage{algorithm}
\usepackage{algpseudocode}

\begin{document}

\title{
Probabilistic Neural Networks (PNNs) with t-Distributed Outputs: Adaptive Prediction Intervals Beyond Gaussian Assumptions%\footnote{This paper is accepted for publication in 2019 IEEE International Workshop on Machine Learning for Signal Processing (MLSP), Pittsburgh, PA, USA.}
}

\author{Farhad Pourkamali-Anaraki\\Department of Mathematical and Statistical Sciences, University of Colorado Denver, CO, USA\\ Email: farhad.pourkamali@ucdenver.edu}

\date{Accepted for Publication in Neural Computing and Applications}
%\date{December 2015}

\maketitle

\abstract{Traditional neural network regression models provide only point estimates, failing to capture predictive uncertainty. Probabilistic neural networks (PNNs) address this limitation by producing output distributions, enabling the construction of prediction intervals. However, the common assumption of Gaussian output distributions often results in overly wide intervals, particularly in the presence of outliers or deviations from normality. To enhance the adaptability of PNNs, we propose t-Distributed Neural Networks (TDistNNs), which generate t-distributed outputs, parameterized by location, scale, and degrees of freedom. The degrees of freedom parameter allows TDistNNs to model heavy-tailed predictive distributions, improving robustness to non-Gaussian data and enabling more adaptive uncertainty quantification. We incorporate a likelihood based on the t-distribution into neural network training and derive efficient gradient computations for seamless integration into deep learning frameworks. Empirical evaluations on synthetic and real-world data demonstrate that TDistNNs  improve the balance between coverage and interval width. Notably, for identical architectures, TDistNNs consistently produce narrower prediction intervals than Gaussian-based PNNs while maintaining proper coverage. This work contributes a flexible framework for uncertainty estimation in neural networks tasked with regression, particularly suited to settings involving complex output distributions.}

\maketitle

\section{Introduction}
\label{sec:intro}
Selecting an appropriate loss function is fundamental to machine learning, as it determines how the model quantifies and reduces its prediction errors \cite{wang2020comprehensive,hu2022understanding,akbari2021theoretical,jadon2024comprehensive}. In parametric models, such as neural networks, this translates to optimizing parameters (weights and biases) to achieve that minimization \cite{pourkamali2021neural}. For regression problems, mean squared error (MSE) is a widely used loss function, calculating the average of the squared differences between predicted and actual output values \cite{fan2025ssim}. This selection profoundly impacts the model's learning behavior, directly influencing its capacity to accurately capture underlying data patterns. Consequently, a poorly chosen loss function can lead to suboptimal performance, even with a well-designed model architecture.

Many loss functions in machine learning are rooted in probabilistic frameworks that model the distribution of predicted values \cite{murphy2022probabilistic}. For instance, it is common to assume that the model's output follows a Gaussian or normal distribution with an unknown mean and a fixed, known variance, regardless of the input. This assumption allows us to treat model training as a parameter estimation problem within statistical inference. Specifically, maximizing the likelihood of observing the training data under this assumption
is mathematically equivalent to minimizing the negative log-likelihood (NLL), which, in this case, is directly proportional to the MSE \cite{pourkamali2024probabilistic}. 
While this choice of loss function enables us to estimate the conditional mean (point prediction), the inherent fixed variance assumption prevents us from assessing the uncertainty associated with those predictions. This limitation underscores the critical need for alternative loss functions, particularly in high-stakes applications where accurate quantification of prediction uncertainty is paramount \cite{nasrin2024application,pourkamali2023evaluation}.

In response to the limitations of fixed-variance models, researchers have developed probabilistic approaches that estimate both the mean and variance of the target variable, assuming a Gaussian distribution \cite{nix1994estimating,seitzer2022pitfalls,immer2023effective,sluijterman2024optimal}. Within neural networks, this is typically implemented by allocating two output neurons for each target variable: one to predict the mean and the other to predict the variance. This architecture allows the construction of a loss function, often termed variance attenuation loss, which is derived from the NLL principle \cite{basora2025benchmark}. The Gaussian density function is favored due to its analytical convenience and its ability to construct prediction intervals. However, a significant drawback is its sensitivity to outliers and model misspecification. The presence of extreme/unusual values often forces the model to overestimate the variance, attempting to encompass these deviations \cite{zhang2024one}. Consequently, the resulting uncertainty estimates become overly broad and less informative.

Beyond Gaussian-based models, alternative loss functions exist that aim to capture the distribution of predicted outputs without imposing strong distributional assumptions. Quantile regression, utilizing the pinball loss, is a prominent example \cite{meinshausen2006quantile,romano2019conformalized,chung2021beyond}. Unlike methods that focus solely on predicting the mean, quantile regression estimates conditional quantiles, offering a flexible framework for characterizing prediction uncertainty across various parts of the distribution. This approach provides significant advantages, particularly its robustness to deviations from normality and its applicability to a wide range of regression problems. However, while quantiles offer a valuable perspective on uncertainty, they do not inherently capture the full shape of the predictive distribution as a parameterized distribution would. Consequently, this can limit the depth of insight into the underlying data generating process.

This paper aims to enhance the precision and calibration of uncertainty estimates in regression by introducing a probabilistic neural network (PNN) framework that utilizes the Student's t-distribution \cite{Ahsanullah2014,li2020review}. In contrast to conventional Gaussian-based PNNs, our approach parameterizes the predictive distribution with the t-distribution. This strategic choice provides increased adaptability, particularly through the degrees of freedom parameter. The t-distribution's ability to model heavier tails significantly reduces sensitivity to extreme/unusual values \cite{min2010deep}. Furthermore, its flexibility extends to approximating the Gaussian distribution when degrees of freedom are large, effectively encompassing the Gaussian model as a special case \cite{dotzel2024learning}. This versatility enhances the model's ability to capture a wider range of data characteristics and improves the reliability of prediction uncertainty estimates.

In this work, we begin by outlining the necessary architectural modifications to implement TDistNNs, our proposed PNNs with t-distributed outputs. Specifically, we transform a deterministic neural network into a probabilistic one by augmenting the output layer. This augmentation involves extending the output from a single mean prediction to the prediction of three parameters: mean, scale, and degrees of freedom, which define the Student's t-distribution. Each parameter is represented by a dedicated neuron in the output layer. We then derive a NLL function, specifically designed to optimize the parameters of the predictive t-distribution.

The analytical derivation of gradients for the custom-designed loss function, with respect to the mean, scale, and degrees of freedom, is another core component of this paper. This approach ensures efficient training via backpropagation and seamless compatibility with popular deep learning libraries, such as PyTorch. In addition, we illustrate how the obtained Student's t-distribution for a given testing point enables the computation of prediction intervals for a user-specified error rate $\alpha$, such as $\alpha=0.1$.
This allows users to obtain prediction intervals where the true value is expected to be contained in $(1-\alpha)$ of cases, based on a single trained neural network.

In summary, we make the following contributions in this work.
\begin{itemize}
\item This paper introduces a novel PNN framework designed to transform traditional deterministic neural networks into models capable of generating comprehensive predictive distributions. Specifically, we propose the utilization of the t-distribution as the foundational statistical model for these predictive distributions, enabling the simultaneous modeling of point predictions, their variability, and the output's tail behavior. This approach extends and generalizes prior research that predominantly relied on Gaussian distributions, offering enhanced adaptability.
\item Building upon the Student's t-distribution framework, we proceed to derive the corresponding NLL loss function for model fitting. This derivation is followed by the analytical computation of the gradients of this loss function with respect to the parameters of the t-distribution, namely the mean, scale, and degrees of freedom. These analytically derived gradients are essential for enabling efficient training via backpropagation, and thus facilitate seamless integration with deep learning frameworks. 
\item To evaluate the performance of the proposed probabilistic framework, we conduct a series of comprehensive experiments involving a detailed comparative analysis with existing methodologies. These include approaches based on parameterized output distributions, such as Gaussian probabilistic neural networks, neural networks trained with the pinball loss for quantile regression, as well as Monte Carlo Dropout-based uncertainty estimation methods. Through these experiments, we analyze the quality of the generated prediction intervals, with a specific focus on their widths and coverage levels. This analysis allows us to thoroughly investigate and present the inherent trade-offs between interval width and coverage accuracy, providing valuable insights into the practical applicability and limitations of the proposed framework and existing approaches.
\end{itemize}

The remainder of this paper is structured as follows. Section \ref{sec:review} reviews relevant notation, existing Gaussian-based PNNs, pinball loss for quantile regression, and Monte Carlo Dropout. Section \ref{sec:proposed} details our proposed model, which extends existing work by using the Student's t-distribution as the predictive distribution, and provides the loss function derivation and gradient computations. In Section \ref{sec:exper}, we present numerical experiments that assess and compare  the quality of prediction intervals. Finally, Section \ref{sec:conclusion} summarizes the key findings, advantages, and future research directions.  

\section{Notation and Literature Review}
\label{sec:review}
We consider a regression problem where $D$-dimensional input features $\mathbf{x} \in \mathbb{R}^D$ are mapped to scalar outputs $y \in \mathbb{R}$. The overall objective is to learn a (stochastic) function $f(\mathbf{x}; \boldsymbol{\theta})$ that accurately models this relationship, given a labeled training data set $\mathcal{D} = \{(\mathbf{x}_1, y_1), \ldots, (\mathbf{x}_N, y_N)\}$. The model parameters $\boldsymbol{\theta}$ are optimized by minimizing a loss function that measures the discrepancy between predicted values $f(\mathbf{x}_n; \boldsymbol{\theta})$ and their corresponding ground-truth outputs $y_n$, for $n=1,\ldots, N$. The mean squared error (MSE) is a widely used loss function for regression, defined as:
\begin{equation}
\ell_{\text{MSE}}(\boldsymbol{\theta}) = \frac{1}{N} \sum_{n=1}^N \big(y_n - f(\mathbf{x}_n; \boldsymbol{\theta})\big)^2.\label{eq:MSE}
\end{equation}
While MSE is computationally efficient and easily interpretable, it is inherently designed for point predictions and does not account for prediction uncertainty. Consequently, a model with high uncertainty is not penalized as long as its point prediction remains close to the true value.

\subsection{Gaussian Output Distribution}

To elucidate the behavior of the MSE loss function, we demonstrate its connection to a probabilistic framework grounded in the Gaussian output distribution. In probabilistic regression, instead of modeling a deterministic function \( f(\mathbf{x}; \boldsymbol{\theta}) \) as a point estimator, we assume that the target variable \( y \) follows a Gaussian distribution \cite{stirn2023faithful}:

\begin{equation}
    p(y \mid \mathbf{x};\boldsymbol{\theta}) \sim \mathcal{N} \big(f_{\mu}(\mathbf{x}; \boldsymbol{\theta}), f_{\sigma}^2(\mathbf{x}; \boldsymbol{\theta})\big), \label{eq:normalOut}
\end{equation}
where \( f_{\mu}(\mathbf{x}; \boldsymbol{\theta}) \) represents the predicted mean and \( f_{\sigma}^2(\mathbf{x}; \boldsymbol{\theta}) \) represents the predicted variance. This formulation allows the model to express both a point prediction and its associated uncertainty.

Given the training set, we estimate the parameters \( \boldsymbol{\theta} \) in Eq.~\eqref{eq:normalOut} by maximizing the likelihood of the observed data. The log-likelihood function for a single observation is \cite{lind2024uncertainty}:
\begin{equation}
    \log p(y_n \mid \mathbf{x}_n; \boldsymbol{\theta}) = -\frac{1}{2} \log (2\pi f_{\sigma}^2(\mathbf{x}_n; \boldsymbol{\theta})) - \frac{(y_n - f_{\mu}(\mathbf{x}_n; \boldsymbol{\theta}))^2}{2 f_{\sigma}^2(\mathbf{x}_n; \boldsymbol{\theta})}.
\end{equation}
Maximizing the log-likelihood is equivalent to minimizing the negative log-likelihood (NLL) function, which leads to the following loss function:

\begin{equation}
\ell_{\text{GaussianNLL}}(\boldsymbol{\theta}) = \frac{1}{N} \sum_{n=1}^{N} \left[ \frac{(y_n - f_{\mu}(\mathbf{x}_n; \boldsymbol{\theta}))^2}{2 f_{\sigma}^2(\mathbf{x}_n; \boldsymbol{\theta})} + \frac{1}{2} \log \big(f_{\sigma}^2(\mathbf{x}_n; \boldsymbol{\theta}) \big)\right].\label{eq:GaussianNLL}
\end{equation}
This formulation provides a crucial benefit. When the variance term \( f_{\sigma}^2(\mathbf{x}_n; \boldsymbol{\theta}) \) is assumed to be fixed and known, the NLL simplifies, leaving only the squared error term \( (y_n - f_{\mu}(\mathbf{x}_n; \boldsymbol{\theta}))^2 \). This reduction makes the loss function equivalent to \(\ell_{\text{MSE}}\) given in Eq.~\eqref{eq:MSE}. Consequently, under this assumption\textemdash known as homoscedasticity\textemdash the training process focuses solely on minimizing the difference between the true and predicted values while disregarding any information about the confidence in model predictions. This limitation highlights the importance of  allowing the model to express varying levels of uncertainty across different inputs.

To further illustrate this point, let us revisit the loss function in Eq.~\eqref{eq:GaussianNLL}. The Gaussian negative log-likelihood (GaussianNLL) loss consists of two key components. 
The first term represents a scaled squared error between the actual output \( y_n \) and the predicted mean \( f_{\mu}(\mathbf{x}_n; \boldsymbol{\theta}) \): \((y_n - f_{\mu}(\mathbf{x}_n; \boldsymbol{\theta}))^2/(2 f_{\sigma}^2(\mathbf{x}_n; \boldsymbol{\theta}))\). This term penalizes deviations of the model's predictions from the true outputs, but unlike the standard MSE loss, the penalty is inversely proportional to the predicted variance. This means that when the model is highly confident (i.e., the predicted variance \( f_{\sigma}^2(\mathbf{x}_n; \boldsymbol{\theta}) \) is small), even slight deviations between \( y_n \) and \( f_{\mu}(\mathbf{x}_n; \boldsymbol{\theta}) \) are penalized more strongly. Conversely, when the predicted variance is large, the model is more tolerant of deviations, as the penalty for prediction errors decreases.

However, relying solely on the first term in Eq.~\eqref{eq:GaussianNLL} would encourage the model to assign arbitrarily large variances to all predictions to minimize the loss. To prevent this, the second term in the GaussianNLL loss acts as a regularization term: \(\frac{1}{2} \log ( f_{\sigma}^2(\mathbf{x}_n; \boldsymbol{\theta}))\). This term discourages excessively large variance predictions by encouraging the model to minimize uncertainty wherever possible. Thus, these two terms work in tandem: the first term ensures that predictions remain close to the observed values, with penalties adjusted based on estimated uncertainty, while the second term prevents the model from arbitrarily inflating variance. We also note that variants of the GaussianNLL loss function exist, such as scaling the loss by a power of the variance \cite{seitzer2022pitfalls} or adopting a Laplace distribution \cite{bramlage2023plausible}.

By training with the GaussianNLL loss function, we can obtain predictions of both the mean and variance, which parameterize the predictive Gaussian distribution during testing. Consider a new feature vector \( \mathbf{x}_{\text{test}} \in \mathbb{R}^D \) and the optimized model parameters \( \boldsymbol{\theta}^* \). The trained model provides estimates for the predictive mean and variance, \( f_{\mu}(\mathbf{x}_{\text{test}}; \boldsymbol{\theta}^*) \) and \( f_{\sigma}^2(\mathbf{x}_{\text{test}}; \boldsymbol{\theta}^*) \), respectively. These estimates facilitate the construction of prediction intervals, which quantify the model's uncertainty about $y_{\text{test}}$ \cite{dewolf2023valid}. Assuming a Gaussian likelihood, the $ (1 - \alpha) \times 100\% $ confidence level prediction interval for the (unknown) true output $y_{\text{test}}$ is defined as:

\begin{equation}
    \left[ f_{\mu}(\mathbf{x}_{\text{test}}; \boldsymbol{\theta}^*) - z_{\alpha/2} \cdot f_{\sigma}(\mathbf{x}_{\text{test}}; \boldsymbol{\theta}^*), 
           f_{\mu}(\mathbf{x}_{\text{test}}; \boldsymbol{\theta}^*) + z_{\alpha/2}\cdot f_{\sigma}(\mathbf{x}_{\text{test}}; \boldsymbol{\theta}^*) \right],\label{eq:pred_inter_normal}
\end{equation}
where $\alpha\in(0,1)$ is the user-specified error rate and \( z_{\alpha/2} \) is the critical value corresponding to the upper tail probability \( \alpha/2 \) in the standard normal distribution. For example, to generate a 90\% confidence level (\( \alpha = 0.1 \)), we use \( z_{0.05} \approx 1.645 \) to calculate the interval's lower and upper bounds.

Modeling the predicted outputs using a Gaussian distribution, as in Eq.~\eqref{eq:normalOut}, provides several advantages. Notably, the predictive mean naturally serves as the point estimate, while the variance quantifies the model's confidence in its predictions \cite{tan2023single}. However, this approach imposes a significant restriction due to its fixed tail behavior, which limits its flexibility in capturing the full range of data distributions. This limitation becomes particularly problematic when dealing with non-Gaussian outputs, such as data sets containing extreme values or rare outputs \cite{lugosi2019mean,ribeiro2020imbalanced,vazquez2025outlier,wibbeke2025quantification}. In such cases, the Gaussian assumption fails to adequately represent the heavier tails often observed in real-world data \cite{zhang2024novel}.  To ensure that prediction intervals remain appropriately calibrated in such complex settings, the model must compensate by inflating the predicted variance. This leads to an overestimation of \( f_{\sigma}^2(\mathbf{x}_{\text{test}}; \boldsymbol{\theta}) \). Consequently, while the Gaussian model maintains nominal coverage, it does so at the cost of unnecessarily broad uncertainty estimates, reducing the informativeness of the prediction intervals.  

\begin{figure}[htbp]
    \centering
\includegraphics[width=0.7\linewidth]{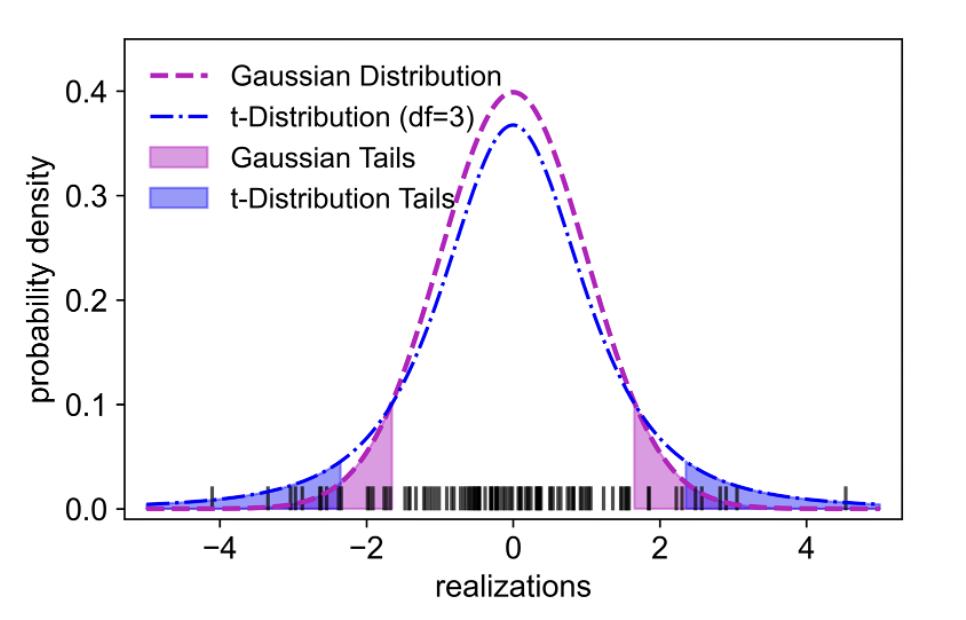}
    \caption{This plot compares the tail behavior of standard Gaussian and t-distributions, showing realizations of a target variable (x-axis) and shaded 0.10 tail probability regions (0.05 per tail). It highlights the t-distribution's heavier tails and its enhanced ability to accommodate extreme values and model mismatch.}
    \label{fig:tails}
\end{figure}

The issue of modeling outliers or extreme values is illustrated in Fig.~\ref{fig:tails}, which compares standard Gaussian and t-distributions, highlighting the regions corresponding to a total tail probability of 0.10 (0.05 on each side). Additionally, it shows a set of realizations of a quantity of interest or output, with the majority clustered around the mean and a few points located in the tails. The Gaussian distribution's light tails fail to adequately capture these extreme points, requiring a much wider variance to encompass them. The t-distribution with a lower degree of freedom, however, with its heavier tails, provides a better fit for the sample data, accommodating rare/extreme events without requiring excessive variance. This underscores the need for enhanced predictive distributions, such as the t-distribution, and their associated loss functions, to provide greater flexibility and robust coverage guarantees for prediction intervals.

\subsection{Quantile Regression}
Another approach to constructing prediction intervals is based on the pinball loss, which is commonly used in quantile regression. Instead of modeling the full distribution of the outputs, as done in Gaussian likelihood-based methods, quantile regression focuses on estimating conditional quantiles of the response variable or output given the input features \cite{steinwart2011estimating}. For a given quantile level \( \tau \in (0,1) \), the pinball loss for a single data point \( (\mathbf{x}_n, y_n) \) is defined as \cite{chung2021beyond}:

\begin{equation}
    \ell_{\text{pinball}}(y_n, f_{\tau}(\mathbf{x}_n; \boldsymbol{\theta})) = 
    \begin{cases}
        \tau (y_n - f_{\tau}(\mathbf{x}_n; \boldsymbol{\theta})), & y_n \geq f_{\tau}(\mathbf{x}_n; \boldsymbol{\theta}) \\
        (1 - \tau) (f_{\tau}(\mathbf{x}_n; \boldsymbol{\theta}) - y_n), & y_n < f_{\tau}(\mathbf{x}_n; \boldsymbol{\theta})
    \end{cases}.\label{eq:pinball}
\end{equation}
This loss function asymmetrically penalizes overestimation and underestimation based on the quantile level \( \tau \). Specifically, when \( y_n \) is greater than the predicted value \( f_{\tau}(\mathbf{x}_n; \boldsymbol{\theta}) \), the loss is scaled by \( \tau \), whereas when \( y_n \) is less than the predicted quantile, the loss is scaled by \( (1 - \tau) \). As a result, the model learns to balance prediction errors in a way that corresponds to the chosen quantile level. 

For example, if we set \( \tau = 0.95 \), the pinball loss penalizes underestimation  much more heavily than overestimation. This is because when the prediction is too low, the residual is multiplied by \( 0.95 \), making the penalty larger, whereas if the prediction is too high, the residual is multiplied by \( 1 - 0.95 = 0.05 \), leading to a much smaller penalty. This behavior encourages the model to adjust predictions upward so that the estimated quantile represents a threshold below which most true values fall. By training separate models for different quantiles, such as \( \tau = 0.05 \) and \( \tau = 0.95 \), one can construct an empirical prediction interval that captures the response variable with an approximate 90\% probability.

However, a key limitation of quantile regression with pinball loss is that it estimates each quantile separately, rather than learning a full predictive distribution in a single process. In contrast, Gaussian likelihood-based methods jointly infer both the mean and variance. Despite this limitation, quantile regression remains a valuable tool for uncertainty estimation, especially in cases where explicit distributional assumptions may not be appropriate.

\subsection{Monte Carlo Dropout}

Monte Carlo Dropout (MC Dropout) is a widely used technique for uncertainty quantification in neural networks that does not require an explicit probabilistic output model \cite{gawlikowski2023survey,he2025survey}. Originally introduced as a regularization strategy during training, dropout operates by  deactivating a subset of neurons to reduce overfitting. This mechanism is controlled by a user-specified dropout rate, which determines the probability with which each neuron is deactivated during a forward pass through the network.

At each training iteration, neurons are stochastically omitted according to this Bernoulli process, while the remaining active units are appropriately rescaled to preserve the expected magnitude of activations. This stochastic architecture perturbation discourages co-adaptation of neurons and promotes the learning of robust, distributed feature representations. As a result, dropout can be viewed as implicitly training an ensemble of subnetworks that share parameters, with each subnetwork corresponding to a particular realization of the dropout mask.

MC Dropout extends this framework to the inference stage by retaining the same stochastic deactivation mechanism during prediction \cite{gal2016dropout}. Rather than producing a single deterministic output for a given input, the model is evaluated multiple times with different dropout realizations, yielding stochastic predictions even for the same input. This induced variability enables the empirical approximation of a predictive distribution through repeated forward passes.

To be more formal, consider a neural network trained using the MSE loss in Eq.~\eqref{eq:MSE}, augmented with dropout layers. At test time, dropout remains active and the model is evaluated $T$ times for a given input $\mathbf{x}_{\text{test}}$, producing a collection of stochastic predictions $
\hat{y}^{(1)}, \hat{y}^{(2)}, \ldots, \hat{y}^{(T)}$,
where $T$ denotes the number of Monte Carlo samples.

These samples define an empirical predictive distribution for the output conditioned on the input $\mathbf{x}_{\text{test}}$. Unlike likelihood-based regression methods, MC Dropout does not explicitly parameterize a conditional distribution such as a Gaussian. Instead, uncertainty is characterized directly from the variability of the Monte Carlo predictions. Consequently, prediction intervals are most naturally constructed using empirical quantiles of the sampled outputs, avoiding additional distributional assumptions. Specifically, a $(1-\alpha)\times100\%$ prediction interval is given by:
\begin{equation}
\left[
Q_{\alpha/2}\bigl(\{\hat{y}^{(t)}\}_{t=1}^{T}\bigr),
\;
Q_{1-\alpha/2}\bigl(\{\hat{y}^{(t)}\}_{t=1}^{T}\bigr)
\right],
\label{eq:mc_PI}
\end{equation}
where $Q_{\tau}(\cdot)$ denotes the empirical $\tau$-quantile of the sampled predictions. This construction naturally accommodates asymmetric or heavy-tailed predictive behavior and is well suited when the shape of the output distribution is unknown.

In summary, MC Dropout provides a practical approach for uncertainty quantification in regression models trained with standard pointwise loss functions. While it enables flexible, assumption-free prediction intervals via sampling, the resulting uncertainty estimates are sensitive to the choice of dropout rate, network architecture, and number of Monte Carlo samples. Moreover, the absence of a likelihood-based training objective limits the principled calibration of predictive uncertainty.

\section{Probabilistic Neural Networks with t-Distributed Outputs}
\label{sec:proposed}
Building upon the GaussianNN framework, this section proposes a modification: replacing the Gaussian assumption for the conditional distribution $p(y\mid \mathbf{x})$ with a Student's t-distribution.
While Gaussian models may provide adequate performance for many regression tasks, they are inherently sensitive to outliers and deviations from normality. The Student's t-distribution, with its heavier tails, offers a more robust alternative, enabling the model to effectively handle extreme values and uncertainties. By introducing the degrees of freedom parameter, the t-distribution provides increased flexibility in capturing the shape of the conditional distribution, leading to more accurate and reliable prediction intervals in complex settings. 

Specifically, we begin by discussing the parameterization of the t-distribution within the PNN framework, detailing how the distribution's location, scale, and shape parameters are derived from the network's outputs. We then address the necessary modifications to standard deterministic neural network architectures to produce the designated predictive t-distribution. Subsequently, we present a detailed derivation of the negative log-likelihood (NLL) loss function, tailored for the t-distribution, and its gradients, which are essential for training the network via backpropagation. %We conclude by detailing how to construct $(1-\alpha)\times 100\%$ confidence level prediction intervals using the learned t-distribution, for any $\alpha$ (e.g., $\alpha=0.1$), thus quantifying prediction uncertainty.

Concretely, we define the predictive distribution in our proposed TDistNN as:
\begin{equation}
    p(y \mid \mathbf{x};\boldsymbol{\theta}) \sim \mathcal{T}\big(f_{\mu}(\mathbf{x}; \boldsymbol{\theta}), f_{\sigma}(\mathbf{x}; \boldsymbol{\theta}), f_{\nu}(\mathbf{x}; \boldsymbol{\theta})\big),
\end{equation}
where $f_{\mu}(\mathbf{x}; \boldsymbol{\theta})\in\mathbb{R}$ denotes the location parameter, $f_{\sigma}(\mathbf{x}; \boldsymbol{\theta})>0$ is the scale parameter, and $f_{\nu}(\mathbf{x}; \boldsymbol{\theta})>0$ is the degrees of freedom (shape) parameter that governs the weight of the tails.
Lower values of $f_{\nu}$
result in heavier tails, enhancing robustness to outliers, while higher values approximate a Gaussian distribution.
  
To keep the notation concise in subsequent derivations, we will write $f_{\mu}$, $f_{\sigma}$, and $f_{\nu}$ without explicitly showing their dependence on the input feature vector $\mathbf{x}$ and the model parameters $\boldsymbol{\theta}$. Nonetheless, it is important to note that these quantities are functions of both the input features and the learnable parameters of the network. 

The probability density function of the t-distribution with the above parameters is given by:
\begin{equation}
    p(y \mid \mathbf{x}; \boldsymbol{\theta}) = \frac{\Gamma\left(\frac{f_{\nu} + 1}{2}\right)}{\sqrt{\pi f_{\nu}} \Gamma\left(\frac{f_{\nu}}{2}\right) f_{\sigma}} \left(1 + \frac{(y - f_{\mu})^2}{f_{\nu} f_{\sigma}^2}\right)^{-\frac{f_{\nu} + 1}{2}},
\end{equation}
where $\Gamma(\cdot)$ is the Gamma function, a continuous generalization of the factorial function to non-integer values. As $f_{\nu} \to \infty$, this distribution converges to a Gaussian, ensuring compatibility with traditional modeling assumptions in GaussianNN. Moreover, when $f_{\nu}>1$, the location parameter $f_{\mu}$ is also the mean of the distribution. 

\begin{figure}[htbp]
    \centering
\includegraphics[width=0.9\linewidth]{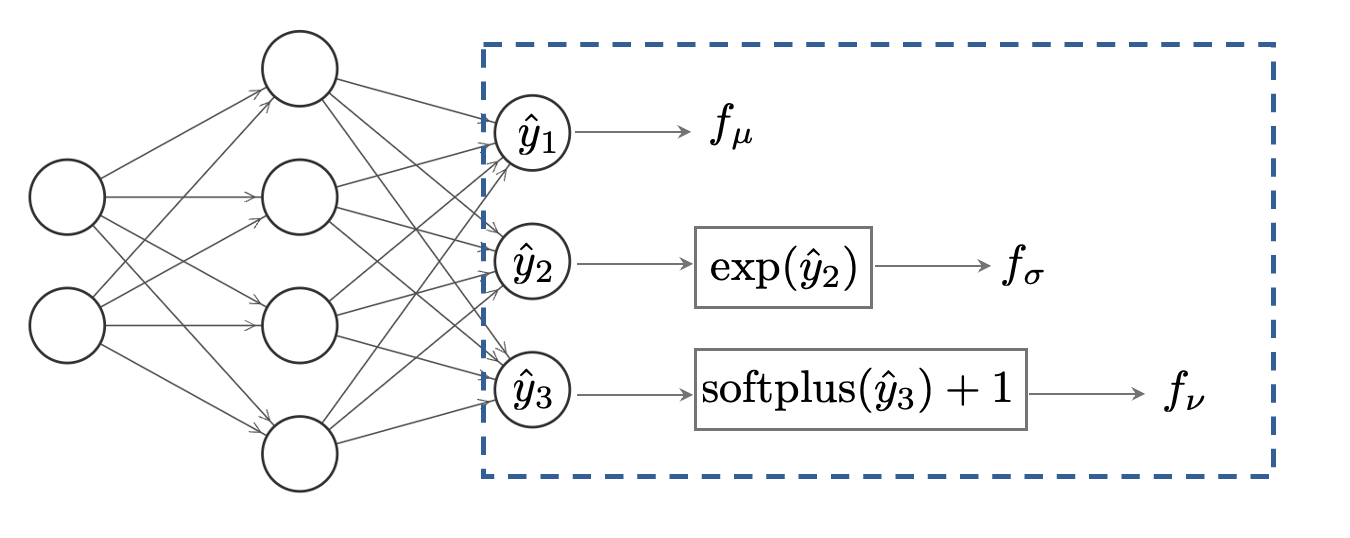}
    \caption{This visualization shows the output layer modifications implemented to define t-Distributed Neural Networks (TDistNNs), ensuring the scale ($f_\sigma$) and degrees of freedom ($f_\nu$) parameters are appropriately constrained, and $f_\mu$ represents the point prediction.}
    \label{fig:tdistnn}
\end{figure}

In a neural network setting, \(f_{\mu}\), \(f_{\sigma}\), and \(f_{\nu}\) can be learned jointly as outputs of the model.
In the case \(f_{\nu} > 1\), \(f_{\mu}\) naturally serves as a point prediction (mean) for the target variable \(y\),
while \(f_{\sigma}\) and \(f_{\nu}\) provide information about the variability and tail behavior of the outputs,
thus capturing uncertainties that go beyond a simple Gaussian assumption. The architecture of the proposed neural networks with t-distributed outputs (TDistNNs) is illustrated in Fig.~\ref{fig:tdistnn}. 
The primary modification to a standard neural network architecture is to include three neurons in the output layer, which we denote by \(\hat{y}_1, \hat{y}_2, \hat{y}_3\). 
Given the discussed constraints in our introduced framework, namely \(f_{\sigma} > 0\) and \(f_{\nu} > 1\), we define the three parameters as follows:
\begin{align}
f_{\mu} &= \hat{y}_1,\nonumber\\
f_{\sigma} &= \exp(\hat{y}_2),\nonumber\\
f_{\nu} &= \mathrm{softplus}(\hat{y}_3) + 1 
= \log\bigl(1 + \exp(\hat{y}_3)\bigr) + 1.
\end{align}

Note that \(\mathrm{softplus}\) \cite{zheng2015improving,pourkamali2024adaptive} is one of the built-in activation functions in many standard deep-learning frameworks. One advantage of \(\mathrm{softplus}\) is its smooth behavior, which avoids the gradient discontinuities encountered in more basic activations such as the Rectified Linear Unit (ReLU) function. 
Additionally, for large negative inputs, \(\mathrm{softplus}\) smoothly saturates near zero, effectively bounding the parameter from below without imposing a hard cutoff. Conversely, for large inputs, 
 \(\mathrm{softplus}\) grows almost linearly, which prevents $f_{\nu}$
  from increasing too rapidly, thereby distinguishing this model from the purely Gaussian case.

We now proceed to compute the NLL function corresponding to the aforementioned t-distribution. To facilitate this, let us consider an input-output pair with index $n$ from the training set $\mathcal{D}$. Hence, we have the following:
\begin{align}
 L_n:=-\log p(y_n \mid \mathbf{x}_n;\boldsymbol{\theta})&=  \frac{1}{2} \log (\pi f_{\nu}) + \log f_{\sigma} - \log \Gamma\left(\frac{f_{\nu} + 1}{2}\right) + \log \Gamma\left(\frac{f_{\nu}}{2}\right)\nonumber \\
    &\quad + \frac{f_{\nu} + 1}{2} \log \left( 1 + \frac{(y_n - f_{\mu})^2}{f_{\nu} f_{\sigma}^2} \right).\label{eq:L_n}
\end{align}

The overall loss function for the proposed t-distribution model, denoted 
\(\ell_{\text{tDistNLL}}(\boldsymbol{\theta})\), is then computed as the average of \(L_n\) across all 
\(N\) training samples:
\begin{align}
  \ell_{\text{tDistNLL}}(\boldsymbol{\theta}) \;=\; \frac{1}{N}\,\sum_{n=1}^{N} L_n.\label{eq:LossT}
\end{align}

 To explain the derived t-distribution loss function, let us first focus on the last term that involves the true output $y_n$ and the predicted location (mean) parameter:
\begin{equation}
\frac{f_{\nu} + 1}{2}
\,\log\Bigl(1 + \frac{(y_n - f_{\mu})^2}{f_{\nu}\,f_{\sigma}^2}\Bigr).
\end{equation}
For large \(f_{\nu}\), the fraction 
\(\tfrac{(y_n - f_{\mu})^2}{f_{\nu}\,f_{\sigma}^2}\)
becomes small, so we use the approximation: \(
\log(1 + z) \;\approx\; z, 
\;
\text{for } z \to 0\).
Hence,
\begin{equation}
\log\Bigl(
   1 + \frac{(y_n - f_{\mu})^2}{f_{\nu}\,f_{\sigma}^2}
\Bigr)
\;\approx\;
\frac{(y_n - f_{\mu})^2}{f_{\nu}\,f_{\sigma}^2}.
\end{equation}
Next, multiply by \(\tfrac{f_{\nu} + 1}{2} \approx \tfrac{f_{\nu}}{2}\) (again valid for large \(f_{\nu}\)):
\[
\frac{f_{\nu} + 1}{2}
\,\frac{(y_n - f_{\mu})^2}{f_{\nu}\,f_{\sigma}^2}
\;\approx\;
\frac{(y_n - f_{\mu})^2}{2\,f_{\sigma}^2},
\]
which is exactly the squared-error part of $\ell_{\text{GaussianNLL}}(\boldsymbol{\theta})$, provided in Eq.~\eqref{eq:GaussianNLL}. Thus, as $f_{\nu}$
  grows, the t-distribution increasingly behaves like a Gaussian model, and its loss encourages $(y_n - f_{\mu})$ to be small in the same way.

  Meanwhile, the other parts of the t-distribution loss function in Eq.~\eqref{eq:LossT} involve factors which do not directly contain the residual, i.e., \(\,(y_n - f_{\mu})\). Instead, they act as regularizers that control both the scale and the shape of the distribution. By assigning an additional cost to large values of \(f_{\sigma}\) and very small or large values of \(f_{\nu}\), these terms help stabilize training and ensure well-behaved parameter estimates. Consequently, the overall loss is meaningful: the final term penalizes the discrepancy between the mean or point prediction and the true output (thanks to the constraint $f_{\nu}>1$), while the remaining components regulate the uncertainty and tail behavior.

To facilitate training of TDistNNs within popular deep learning frameworks, such as PyTorch, 
we now discuss how to compute the partial derivatives of the t-distribution loss function $ \ell_{\text{tDistNLL}}(\boldsymbol{\theta})$ with respect to 
\(f_{\mu}\), \(f_{\sigma}\), and \(f_{\nu}\).
Since the overall loss is simply the average of \(L_n\) over all training samples, 
it suffices to focus on the partial derivatives of each \(L_n\) with respect to these quantities.

To simplify the differentiation, we introduce the following auxiliary variables:
\begin{align}
    r_n &:= y_n \;-\; f_{\mu}, \nonumber \\
    z_n &:= r_n/f_\sigma \nonumber \\
    s_n &:= 1 \;+\;
    z_n^2/f_{\nu}.
\end{align}

\paragraph{Partial derivative with respect to \boldmath\(f_{\mu}\).}
Only the last term of Eq.~\eqref{eq:L_n} depends on \(f_{\mu}\). 
Using the chain rule and noting that \(\tfrac{\partial r_n}{\partial f_{\mu}} = -1\), 
we obtain:
\begin{align}
  \frac{\partial L_n}{\partial f_{\mu}}
  &= \frac{\partial}{\partial f_{\mu}}
       \left[\frac{f_{\nu} + 1}{2} \,\log\bigl(s_n\bigr)\right]
   \;=\; \frac{f_{\nu} + 1}{2} \;\frac{1}{s_n}
         \,\frac{\partial s_n}{\partial f_{\mu}} \nonumber \\
  &=\; \frac{f_{\nu} + 1}{2} \;\frac{1}{s_n} 
         \left[\frac{2\,z_n}{f_{\nu}}\;\frac{\partial z_n}{\partial f_{\mu}}\right] \nonumber \\
  &= \frac{f_{\nu} + 1}{2} \;\frac{1}{s_n} 
     \left[\frac{2\,z_n}{f_{\nu}}\;\Bigl(-\frac{1}{f_{\sigma}}\Bigr)\right] 
   \;=\; -\,\frac{(f_{\nu}+1)\,r_n}{s_n\,f_{\nu}\,f_{\sigma}^2}.\label{eq:mu}
\end{align}
Above, we use \(\tfrac{\partial z_n}{\partial f_{\mu}} 
= \tfrac{\partial}{\partial f_{\mu}}\bigl(\tfrac{r_n}{f_{\sigma}}\bigr)
= -\tfrac{1}{f_{\sigma}}\).

\paragraph{Partial derivative with respect to \boldmath\(f_{\sigma}\).}
Two terms in Eq.~\eqref{eq:L_n} depend on \(f_{\sigma}\): 
(1)~the \(\log(f_{\sigma})\) term and 
(2)~the \(\log(s_n)\) term. 
Hence, we get:
\begin{align}
  \frac{\partial L_n}{\partial f_{\sigma}}
  &= \frac{\partial}{\partial f_{\sigma}}
     \Bigl[\log\bigl(f_{\sigma}\bigr)\Bigr]
     \;+\; \frac{\partial}{\partial f_{\sigma}}
     \Bigl[\frac{f_{\nu} + 1}{2}\,\log\bigl(s_n\bigr)\Bigr].
\end{align}
The first derivative is simply \(1/f_{\sigma}\). 
For the second, note that 
\(\tfrac{\partial s_n}{\partial f_{\sigma}}
= \tfrac{\partial}{\partial f_{\sigma}}\Bigl(1 + \tfrac{z_n^2}{f_{\nu}}\Bigr)
= \tfrac{1}{f_{\nu}}\,2\,z_n\,\tfrac{\partial z_n}{\partial f_{\sigma}}\).
Since \(z_n = \tfrac{r_n}{f_{\sigma}}\), 
we have 
\(\tfrac{\partial z_n}{\partial f_{\sigma}} 
= \tfrac{\partial}{\partial f_{\sigma}}\Bigl(\tfrac{r_n}{f_{\sigma}}\Bigr)
= -\,\tfrac{r_n}{f_{\sigma}^2}\).
Putting it all together:
\begin{align}
  \frac{\partial L_n}{\partial f_{\sigma}}
  &= \; \frac{1}{f_{\sigma}}
    \;+\; 
     \frac{f_{\nu}+1}{2} \;\frac{1}{s_n} 
     \left[\frac{2\,z_n}{f_{\nu}}
           \left(-\frac{r_n}{f_{\sigma}^2}\right)\right] \nonumber \\
  &= \frac{1}{f_{\sigma}}
    \;-\; 
     \frac{(f_{\nu}+1)\,r_n^2}{s_n\,f_{\nu}\,f_{\sigma}^3}.\label{eq:sigma}
\end{align}

\paragraph{Partial derivative with respect to \boldmath\(f_{\nu}\).}
Several terms in Eq.~\eqref{eq:L_n} depend on \(f_{\nu}\):
\(\frac{1}{2}\,\log(\pi f_{\nu})\), 
\(\log\Gamma(\tfrac{f_{\nu}}{2})\), 
\(\log\Gamma(\tfrac{f_{\nu}+1}{2})\), 
and the final \(\log(s_n)\) term. 
We recall that the derivative of 
\(\log\Gamma(z)\) with respect to \(z\) is the digamma function \(\psi(z)\) \cite{bernardo1976psi,yin2025some}. 
Therefore, we have:
\begin{align}
  \frac{\partial L_n}{\partial f_{\nu}}
  &= \frac{1}{2}\,\frac{\partial}{\partial f_{\nu}} \bigl[\log(\pi f_{\nu})\bigr]
   \;-\;\frac{\partial}{\partial f_{\nu}}
         \bigl[\log \Gamma\bigl(\tfrac{f_{\nu}+1}{2}\bigr)\bigr]
   \;+\;\frac{\partial}{\partial f_{\nu}}
         \bigl[\log \Gamma\bigl(\tfrac{f_{\nu}}{2}\bigr)\bigr] \nonumber\\
  &\quad
   + \frac{\partial}{\partial f_{\nu}}
     \left[\frac{f_{\nu} + 1}{2}\,\log\bigl(s_n\bigr)\right].
\end{align}
We address each term:

\begin{itemize}
\item 
\(\tfrac{\partial}{\partial f_{\nu}} \bigl[\tfrac{1}{2}\,\log(\pi f_{\nu})\bigr]
= \tfrac{1}{2}\,\frac{1}{f_{\nu}}\).
\item 
\(\tfrac{\partial}{\partial f_{\nu}} \bigl[-\,\log\Gamma\bigl(\tfrac{f_{\nu}+1}{2}\bigr)\bigr]
= -\,\frac{1}{2}\,\psi\!\Bigl(\tfrac{f_{\nu}+1}{2}\Bigr).\)
\item 
\(\tfrac{\partial}{\partial f_{\nu}} \bigl[\log\Gamma\bigl(\tfrac{f_{\nu}}{2}\bigr)\bigr]
= \frac{1}{2}\,\psi\!\Bigl(\tfrac{f_{\nu}}{2}\Bigr).\)
\item 
\(\tfrac{\partial}{\partial f_{\nu}}
     \Bigl[\tfrac{f_{\nu} + 1}{2}\,\log\bigl(s_n\bigr)\Bigr]
\) 
involves a product rule. 
First, 
\(\tfrac{\partial}{\partial f_{\nu}}
     \Bigl[\tfrac{f_{\nu} + 1}{2}\Bigr] 
= \tfrac{1}{2}\). 
Second, 
\(\tfrac{\partial \log(s_n)}{\partial f_{\nu}}
= \frac{1}{s_n}\,\tfrac{\partial}{\partial f_{\nu}}
   \Bigl(1 + \frac{z_n^2}{f_{\nu}}\Bigr)
= \frac{1}{s_n}
   \left(-\,\frac{z_n^2}{f_{\nu}^2}\right).
\)
Putting these together:
\begin{align}
  \tfrac{\partial}{\partial f_{\nu}}
     \Bigl[\tfrac{f_{\nu} + 1}{2}\,\log\bigl(s_n\bigr)\Bigr]
  \;=\;
  \tfrac{1}{2}\,\log\bigl(s_n\bigr)
  \;+\; \tfrac{f_{\nu} + 1}{2}\,\frac{1}{s_n}
         \Bigl(-\,\frac{z_n^2}{f_{\nu}^2}\Bigr).
\end{align}
\end{itemize}

Combining all contributions, we obtain the following result:
\begin{align}
  \frac{\partial L_n}{\partial f_{\nu}}
  &= \frac{1}{2\,f_{\nu}}
    \;-\; \frac{1}{2}\,\psi\!\Bigl(\tfrac{f_{\nu}+1}{2}\Bigr)
    \;+\; \frac{1}{2}\,\psi\!\Bigl(\tfrac{f_{\nu}}{2}\Bigr)\nonumber \\&
    \;+\; \frac{1}{2}\,\log\bigl(s_n\bigr)
    \;-\; \frac{(f_{\nu} + 1)\,r_n^2}{2\,s_n\,f_{\nu}^2f_{\sigma}^2}.\label{eq:nu}
\end{align}

In the following, we combine the preceding derivations and illustrate how they fit together to train PNNs with a t-distributed output. For each training data point \((\mathbf{x}_n, y_n)\) from $\mathcal{D}$, one evaluates the forward pass to obtain \(\bigl(f_{\mu},f_{\sigma},f_{\nu}\bigr)\), computes the t-distribution NLL,
\begin{equation}
   L_n \;=\; -\log p(y_n \mid \mathbf{x}_n; f_{\mu}, f_{\sigma}, f_{\nu}),
\end{equation}
then uses the derived partial derivatives to backpropagate and update the network parameters \(\boldsymbol{\theta}\). By repeating this procedure in mini-batches or over the entire training set using a standard gradient-based optimizer (e.g., SGD or Adam \cite{kingma2014adam}), we obtain:
\begin{equation}
\boldsymbol{\theta}^{\ast}\;\in\;\underset{\boldsymbol{\theta}}{\arg\min}\;\ell_{\text{tDistNLL}}(\boldsymbol{\theta}),
\end{equation}
where $\boldsymbol{\theta}^{\ast}$ are the optimized model parameters that map
inputs \(\mathbf{x}\) to the t-distribution parameters \(\bigl(f_{\mu}, f_{\sigma}, f_{\nu}\bigr)\).

After training, the network provides point predictions via \(f_{\mu}(\mathbf{x}_{\text{test}}; \boldsymbol{\theta}^{\ast})\) for a new testing point $\mathbf{x}_{\text{test}}$,  and quantifies uncertainty using the scale \(f_{\sigma}(\mathbf{x}_{\text{test}}; \boldsymbol{\theta}^{\ast})\) and degrees of freedom \(f_{\nu}(\mathbf{x}_{\text{test}}; \boldsymbol{\theta}^{\ast})\). Crucially, to construct a \((1 - \alpha)\times 100\%\) confidence level prediction interval for \(\mathbf{x}_{\mathrm{test}}\), one leverages the ``critical values'' of the t-distribution. If $t_{\alpha/2}:=t_{\alpha/2}(f_{\nu}(\mathbf{x}_{\mathrm{new}};\boldsymbol{\theta}^{\ast}))$
denotes the critical value corresponding to the upper tail probability $\alpha/2$ in the t-distribution with the shape or degrees of freedom parameter $f_{\nu}(\mathbf{x}_{\mathrm{new}};\boldsymbol{\theta}^{\ast})$,  then we get the following prediction interval for $\mathbf{x}_{\text{test}}$ using our proposed TDistNN:
\begin{equation}
\Bigl[
f_{\mu}(\mathbf{x}_{\mathrm{new}};\boldsymbol{\theta}^{\ast}) 
\;-\; t_{\alpha/2} \cdot 
       f_{\sigma}(\mathbf{x}_{\mathrm{new}};\boldsymbol{\theta}^{\ast}),\;
f_{\mu}(\mathbf{x}_{\mathrm{new}};\boldsymbol{\theta}^{\ast}) 
\;+\; t_{\alpha/2} \cdot
       f_{\sigma}(\mathbf{x}_{\mathrm{new}};\boldsymbol{\theta}^{\ast})
\Bigr].
\end{equation}
Compared to the Gaussian-based prediction interval in Eq.~\eqref{eq:pred_inter_normal}, the primary distinction in this equation is that $t_{\alpha/2}$
  provides finer control over the tail behavior of the predictive distribution, thereby influencing the associated probabilities. 
  As such, TDistNN provides both a robust point estimate and reliable predictive intervals\textemdash particularly advantageous in the presence of outliers or heavy-tailed data. The overall procedure is summarized in Alg.~\ref{alg:TDistNN}.

\begin{algorithm}[ht]
\caption{t-Distributed Neural Network (TDistNN)}
\label{alg:TDistNN}
\begin{algorithmic}[1]

\State \textbf{Inputs:}
\Comment{Training data and hyperparameters}
\Statex \quad \(\mathcal{D} = \{(\mathbf{x}_n, y_n)\}_{n=1}^N\): training data set
\Statex \quad Network architecture: number of hidden layers, activation functions, etc.
\Statex \quad \(\eta\): learning rate (or schedule)
\Statex \quad \text{Choice of optimizer (e.g., SGD, Adam)}
\Statex \quad \(\alpha\in(0,1)\): error rate for constructing prediction intervals (if desired)

\vspace{1mm}
\State \textbf{Outputs:}
\Comment{Optimal weights and predictive mapping}
\Statex \quad \(\boldsymbol{\theta}^{\ast}\): optimal network weights/biases
\Statex \quad Trained mapping \(\mathbf{x} \mapsto \bigl(f_{\mu}(\mathbf{x}; \boldsymbol{\theta}^{\ast}), 
                                             f_{\sigma}(\mathbf{x}; \boldsymbol{\theta}^{\ast}), 
                                             f_{\nu}(\mathbf{x}; \boldsymbol{\theta}^{\ast})\bigr)\)
\Statex \quad  Prediction interval with $(1-\alpha)$ coverage (optional)

\vspace{1mm}
\State \textbf{Procedure:}
\State \(\boldsymbol{\theta} \leftarrow \text{Initialize randomly}\)
\Repeat
    \State Sample a mini-batch \(\{(\mathbf{x}_b, y_b)\}\subset\mathcal{D}\)
    \ForAll{ \((\mathbf{x}_b, y_b)\) in mini-batch}
        \State \textbf{Forward pass:}
        \Statex \quad\quad\quad \((\hat{y}_1,\hat{y}_2,\hat{y}_3) 
          = \text{ForwardPass}(\mathbf{x}_b; \boldsymbol{\theta})\)
        \Statex \quad \quad \quad \(f_{\mu}=\hat{y}_1,\; f_{\sigma}=\exp(\hat{y}_2),\;
                 f_{\nu}=\log\bigl(1+\exp(\hat{y}_3))+1\)
        \State \textbf{Compute NLL:}
        \Statex \quad \quad \quad  \(\displaystyle
            L_b \;=\; -\log p\bigl(y_b \mid \mathbf{x}_b; f_{\mu}, f_{\sigma}, f_{\nu}\bigr)\)
    \EndFor
    \State \(\displaystyle
       \ell_{\text{tDistNLL}}(\boldsymbol{\theta}) 
       = \frac{1}{|\text{mini-batch}|}\,\sum L_b
     \)
    \State \textbf{Backpropagate:} 
       \(\displaystyle \frac{\partial \ell_{\text{tDistNLL}}}{\partial \boldsymbol{\theta}}
         \;\leftarrow\;\text{via chain rule w.r.t.\ } f_{\mu}, f_{\sigma}, f_{\nu}\) \Comment{Equations \eqref{eq:mu}, \eqref{eq:sigma}, \eqref{eq:nu}}
    \State \textbf{Update parameters:}
       \(\displaystyle \boldsymbol{\theta} 
        \;\leftarrow\; \boldsymbol{\theta}
        \;-\;\eta \;\frac{\partial \ell_{\text{tDistNLL}}}{\partial \boldsymbol{\theta}}\) \Comment{Basic update formula}
\Until{convergence}

\State \textbf{Result:} \(\boldsymbol{\theta}^{\ast}\) and final TDistNN mapping.

\vspace{1mm}
\State \textbf{(Optional) prediction interval for a new input \(\mathbf{x}_{\mathrm{test}}\):}
\Statex \quad \(\bigl(f_{\mu}, f_{\sigma}, f_{\nu}\bigr) 
         = \text{TDistNN}\bigl(\mathbf{x}_{\mathrm{test}}; \boldsymbol{\theta}^{\ast}\bigr)\).
\Statex \quad Let $t_{\alpha/2}:=t_{\alpha/2}(f_{\nu})$ be the critical value. Then the prediction interval is: \(\displaystyle
          f_{\mu} \pm t_{\alpha/2}
f_{\sigma}
        \)

\end{algorithmic}
\end{algorithm}

\section{Experimental Results}\label{sec:exper}
In this section, we evaluate the performance of TDistNNs through a systematic comparison with existing methods. Our focus is on the construction of prediction intervals, a fundamental approach to quantifying uncertainty in regression tasks. Unlike point estimates, prediction intervals offer a probabilistic measure of confidence, enhancing model interpretability and ensuring more reliable decision-making in real-world applications where uncertainty is inherent, such as solving design optimization problems \cite{pourkamali2024two}. As detailed in the previous section, TDistNNs generate a point estimate via their first output node (see Fig.~\ref{fig:tdistnn}), while the scale and shape parameters of the fitted t-distribution define the interval bounds. This structure facilitates a direct and rigorous comparison with Gaussian-based PNNs (GaussianNN), quantile regression models optimized using the pinball loss (QuantileNN), and Monte Carlo Dropout (MC Dropout).

To assess performance comprehensively, we focus on two critical evaluation metrics for a test data set: (1) the coverage score, expressed as a percentage of true values contained within the prediction intervals, and (2) the interval width, defined as the average width of the predicted intervals across all test samples. A narrower interval indicates higher precision, while a wider interval suggests greater uncertainty. The ideal model strikes a balance, maintaining high coverage while keeping interval width as tight as possible. In all experiments, we form 90\% confidence level prediction intervals by setting the error rate parameter $\alpha=0.1$.  Consequently, a better-performing model is one that meets or exceeds the desired coverage while minimizing the interval width.

A key objective of this paper is to show that integrating TDistNN into an existing neural network architecture is both straightforward and comparable to integrating other methods. To ensure a fair comparison, we minimize specialized hyperparameter tuning and maintain a consistent architecture throughout our analysis. Accordingly, we use a single-hidden-layer neural network for all experiments in this section, except for the final experiment, which considers deeper architectures.

We implement the proposed TDistNNs in PyTorch, which is a powerful tool for building and training neural networks. Conveniently, PyTorch has a built-in function \texttt{lgamma}, which computes the logarithm of the Gamma function. This function is essential for computing $L_n$ in Eq.~\eqref{eq:L_n}. Similarly, PyTorch includes a built-in loss function for the Gaussian negative log-likelihood, named \texttt{GaussianNLLLoss}, which facilitates training Gaussian-based probabilistic models. A custom loss function was developed to implement the pinball loss function for training QuantileNNs.

Unless stated otherwise, all models are trained using the Adam optimizer with a learning rate of 0.01 for 1,000 epochs, ReLU activation in the hidden layer, and a batch size corresponding to the entire training data set. Each method employs output-layer activations tailored to its respective distribution. For TDistNNs, we use the exponential function for the scale parameter and softplus for the shape parameter, ensuring valid t-distribution parameters. Similarly, in GaussianNNs, the exponential function is used to predict the variance of the output normal distribution. For QuantileNNs, we apply a standard linear activation function in the output layer. However, as mentioned, QuantileNNs require two separate training runs to estimate the desired prediction interval, corresponding to the quantile levels $\tau_1 = 0.05$ and $\tau_2 = 0.95$. For MC Dropout, we use a dropout rate of 0.2 with an ensemble size of $T=100$. 

\subsection{Synthetic Data with Heteroscedastic Noise and Outliers}

One of the key motivations for developing neural networks with t-distributed outputs is their ability to provide reliable prediction intervals when the Gaussian assumption is violated, particularly in the presence of outliers. To investigate this, our first experiment utilizes a synthetically generated data set to evaluate TDistNNs in a controlled setting. We simulate a heteroscedastic regression problem where the true relationship follows: \(  y_n = 2 + 3x_n + \epsilon_n\). Here, the scalar input $x_n \in [0,5]$ and the noise term \(\epsilon_n\) is drawn from a zero-mean Gaussian distribution with a standard deviation of \(0.5x_n\), making the noise level input-dependent. We generate 1,000 training samples from this model and introduce outliers by selecting 10\% of the data points and adding additional noise drawn from a zero-mean Gaussian distribution with three times the base standard deviation. This process injects extreme values into the training set, allowing us to assess how well TDistNNs handle heavy-tailed noise and deviations from normality.

\begin{figure}[htbp]
    \centering
\includegraphics[width=\linewidth]{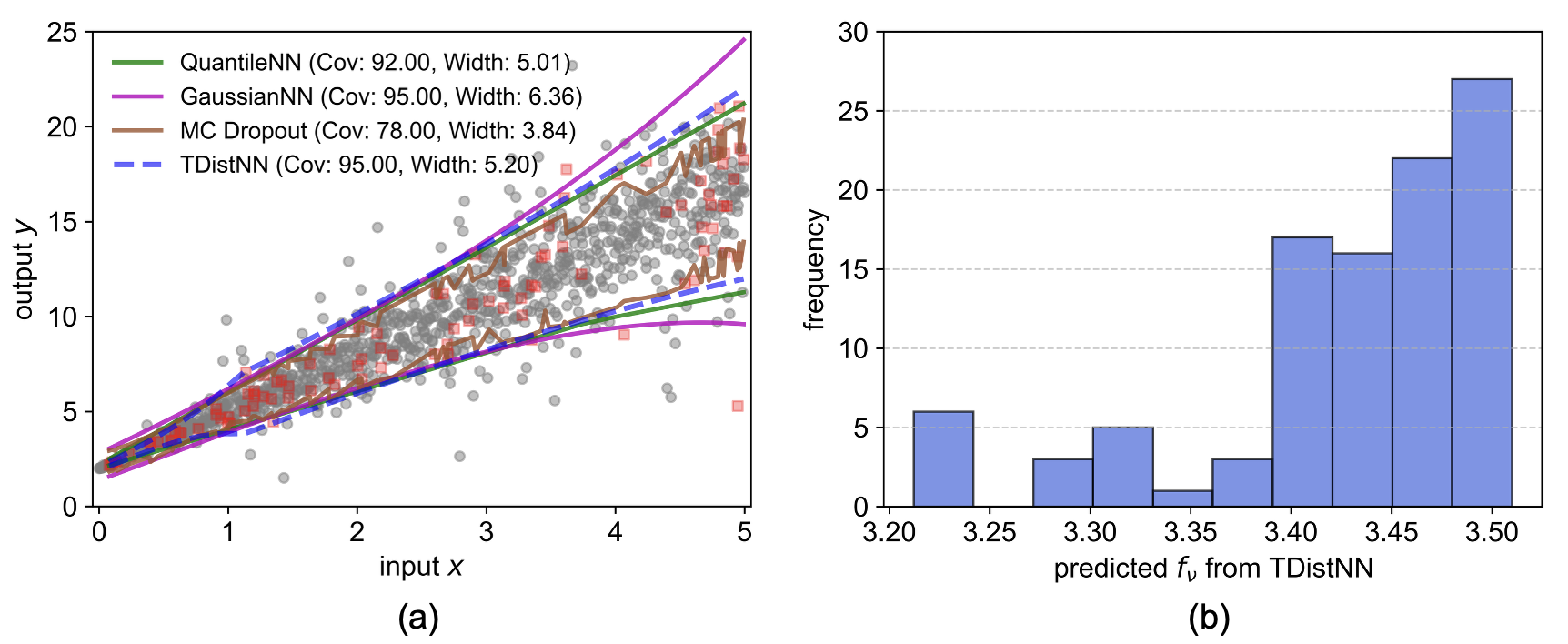}
    \caption{Using a synthetic data set with outliers, we compare the performance of our proposed TDistNN against QuantileNN, GaussianNN, and MC Dropout. (a) shows the constructed prediction intervals, and (b) displays the histogram plot of degrees of freedom $f_{\nu}$ of the fitted t-distribution.}
    \label{fig:syn-bounds}
\end{figure}

The base neural network model utilizes a single hidden layer with 16 units. To evaluate performance, we generate 100 test samples using the same data generation process as the training set. Fig.~\ref{fig:syn-bounds}(a) visualizes the constructed prediction intervals for these test points, illustrating the model's adaptability across the input space. We also report the coverage scores and average interval widths for the three methods under investigation.

We observe that the QuantileNN model produces the narrowest prediction intervals throughout the entire input range while maintaining the target 90\% coverage (recall that we set $\alpha=0.1$). While MC Dropout produces narrower intervals, this comes at the cost of substantial undercoverage (78\% versus the target 90\%), and its resulting prediction bounds are noticeably less smooth than those of the other methods. In contrast, both GaussianNN and TDistNN models achieve slightly higher coverage scores, with 95\% of true outputs falling within their respective prediction intervals. However, a significant advantage of the proposed TDistNN model is its reduced average interval width. At 5.20, it is 18.24\% narrower than GaussianNN's 6.36, while maintaining comparable coverage. This improvement is particularly noticeable at the input range's extremities, where GaussianNN's intervals widen to accommodate the outliers. 

Therefore, the experimental results validate the t-distribution's effectiveness in generating adaptive prediction intervals, outperforming the Gaussian model and providing a slight coverage enhancement over the QuantileNN model. Moreover, QuantileNNs necessitate multiple training iterations to approximate the predictive distribution due to its reliance on pre-defined quantiles.

In Fig.~\ref{fig:syn-bounds}(b), we present the empirical distribution of the t-distribution's shape parameter, ($f_{\nu}$), representing the degrees of freedom. As expected, all estimated degrees of freedom are greater than 1. The observed magnitudes of $f_{\nu}$ indicate the presence of heavier tails relative to the Gaussian distribution, facilitating narrower prediction intervals in the TDistNN model while preserving the target coverage level of 90\%. This observation highlights the critical role of explicitly modeling the shape parameter, which directly governs the tail behavior of the predictive distribution. In contrast to Gaussian-based methods that assume a fixed tail decay, TDistNN dynamically adapts to varying levels of uncertainty, particularly in scenarios involving outliers or heteroscedastic noise.

To investigate the influence of network architecture on prediction performance, we systematically vary the number of units within a single hidden layer. Specifically, we evaluate architectures with \{8, 16, 32\} hidden units. It is well-established that neural network training exhibits stochastic behavior, stemming from factors such as random weight initialization and inherent algorithmic variations \cite{picard2021torch,scabini2024improving}. To mitigate the impact of this variability and obtain robust performance estimates, we perform 20 independent training trials for each architecture. For each trial, we record both the coverage score and the average prediction interval width. To visualize and summarize the distribution of these metrics across the 20 trials, we employ boxplots. 

\begin{figure}
    \centering
\includegraphics[width=\linewidth]{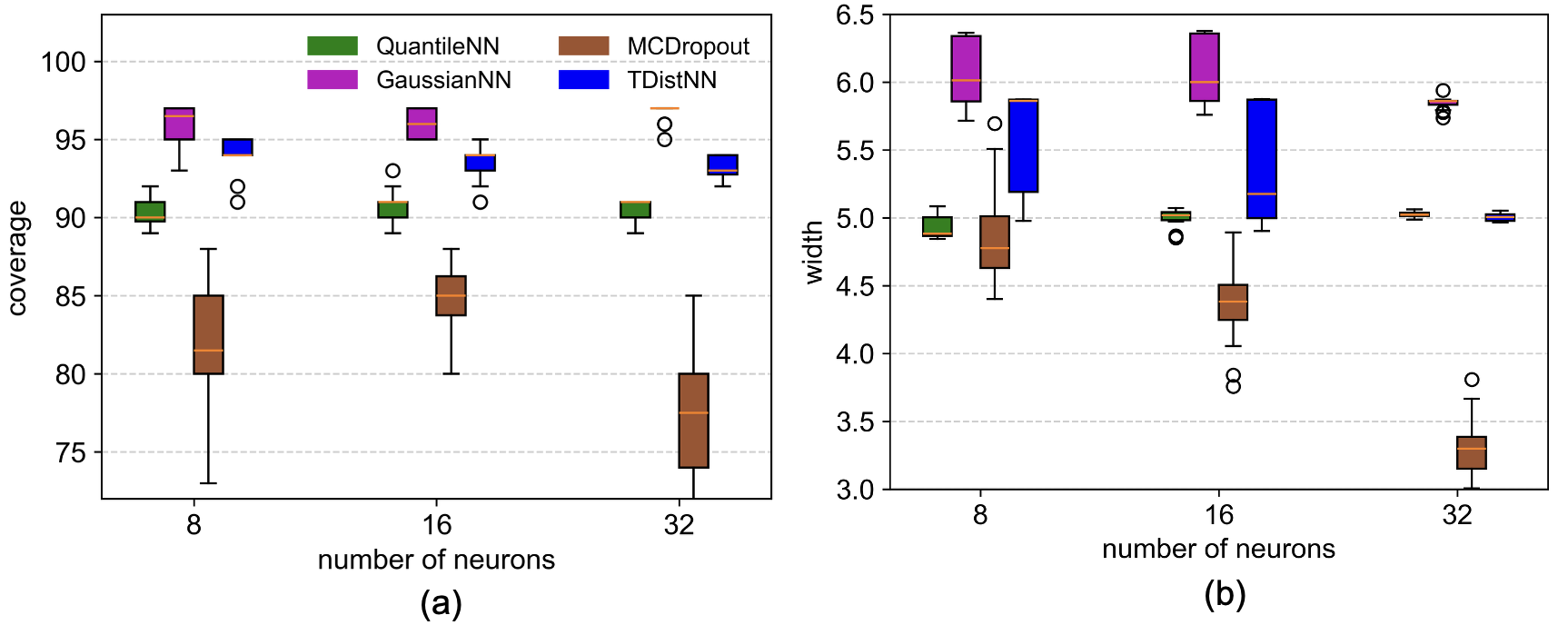}
    \caption{Boxplots illustrate (a) coverage score and (b) average interval width for different architectures across 20 trials on the synthetic data set with outliers. For each network size, the results are shown in the following left-to-right order: QuantileNN, GaussianNN, MC Dropout, and TDistNN.}
    \label{fig:syn_cov_wdi}
\end{figure}

The coverage scores presented in Fig.~\ref{fig:syn_cov_wdi}(a) reveal that GaussianNN and TDistNN consistently surpass the 90\% target coverage. In contrast, MC Dropout fails to reach the 90\% target coverage in any of the 20 trials. While QuantileNN's median coverage is adequate, its minimum values fall below the target coverage across all network architectures. Moreover, a consistent trend emerges: GaussianNNs exhibit higher coverage scores than TDistNNs. To evaluate whether this increased coverage is attributable to wider intervals designed to accommodate outliers within this data set, we must analyze the interval widths depicted in Fig.~\ref{fig:syn_cov_wdi}(b).

The expected trend of GaussianNNs producing significantly wider prediction intervals than TDistNNs is observed, indicating that its higher coverage is a result of inflated uncertainty estimates. TDistNNs, in contrast, effectively construct more adaptive prediction intervals, striking a favorable balance between uncertainty quantification and interval tightness. %Furthermore, a reduction in the variability of interval widths across the 20 independent trials is evident when employing 32 neurons in the hidden layer.
When employing 32 neurons,  both QuantileNN and TDistNN achieve a median interval width of approximately 5, but TDistNN's minimum coverage of 92\% demonstrates a clear advantage in reliability. These results collectively highlight the merits of TDistNNs in complex data settings and elucidate the influence of network architecture on the key evaluation metrics. In addition, we observe that for MC Dropout, when the number of neurons is 32, the variability across the 20 trials in both coverage and interval width is higher than that of the other models.

\subsection{UCI Regression Benchmarks}

The second part of this section evaluates our probabilistic framework with t-distributed outputs on two regression data sets from the UCI Machine Learning Repository: the Concrete Compressive Strength data set \cite{hariri2024benchmarking} (1030 samples, predicting concrete strength, with output values spanning 2.33 to 82.6) and the Energy Efficiency data set \cite{tsanas2012accurate} (768 samples, predicting energy loads, with output values spanning 6.01 to 43.10), both data sets containing 8 input features. We employ a standard 80/20 train-test split, reserving 20\% of the data for evaluating prediction interval quality. Due to the moderate sample sizes, we focus on base neural networks with a single hidden layer, varying the number of neurons within the set \{8, 16, 32\}. This corresponds to hidden layer sizes that are 1, 2, and 4 times the input layer size. To mitigate the effects of neural network training stochasticity, each experiment was conducted 20 times to present boxplots. 

\begin{figure}[htbp]
    \centering
    \includegraphics[width=\linewidth]{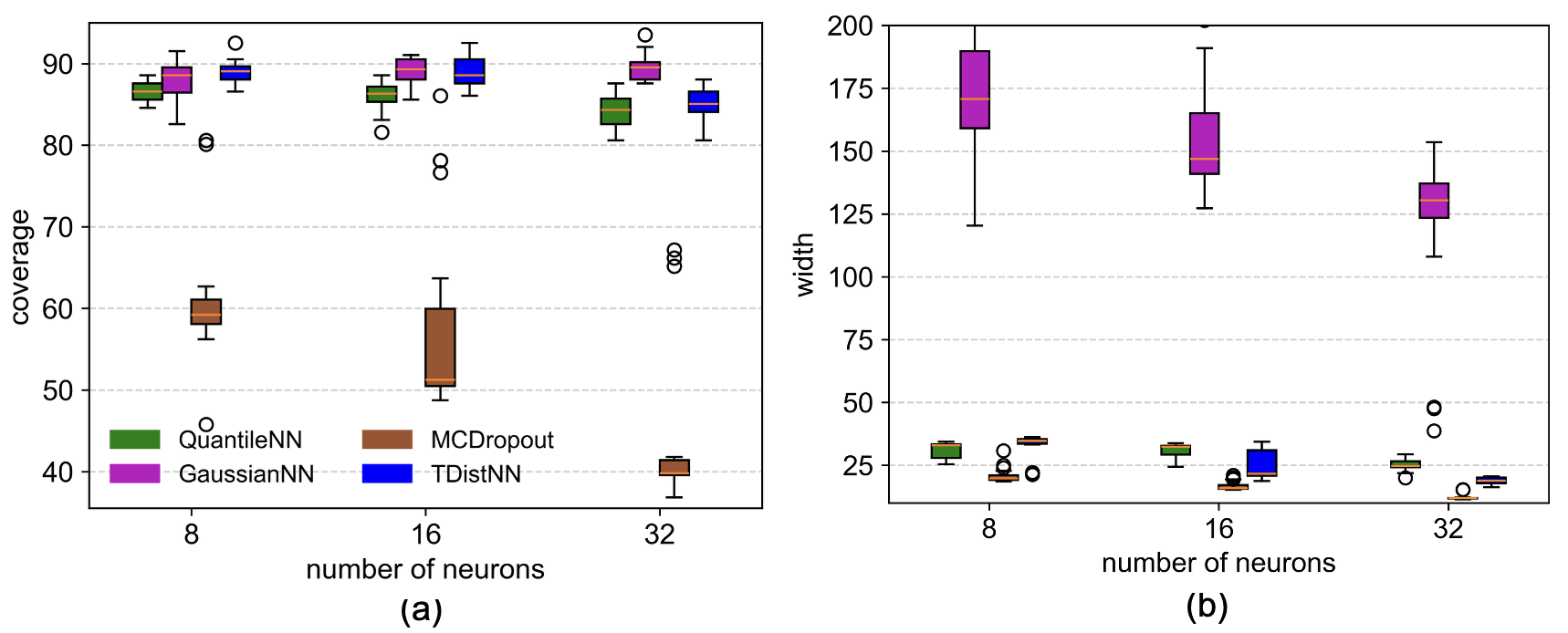}
    \caption{Boxplots illustrate (a) coverage score and (b) interval width for different architectures across 20 trials on the Concrete Compressive Strength data set. For each network size, the results are shown in the following left-to-right order: QuantileNN, GaussianNN, MC Dropout, and TDistNN.}
    \label{fig:concrete_cov}
\end{figure}

We begin by presenting the results for the Concrete Compressive Strength data set, visualized in Fig.~\ref{fig:concrete_cov}. Examining the coverage scores in Fig.~\ref{fig:concrete_cov}(a), it is evident that none of the evaluated methods consistently achieve the target coverage level, indicating the inherent complexity and distributional challenges of this data set. This suggests that the Concrete Compressive Strength data deviates significantly from simple distributional assumptions, making accurate prediction interval estimation difficult. Focusing on coverage scores, TDistNNs demonstrate the highest coverage with 8 hidden neurons, and also achieve a superior maximum coverage score compared to GaussianNNs with 16 neurons, while GaussianNNs excel with 32 neurons. Notably, QuantileNNs consistently underperform, failing to reach the 90\% coverage level across all configurations. As in the previous experiment, MC Dropout suffers from severe undercoverage in this case study, with median coverage ranging between 40\% and 60\%, well below the target coverage of 90\%.

A stark contrast emerges when analyzing average prediction interval widths across 20 trials. GaussianNNs consistently generate excessively wide intervals (medians: 170.77, 146.84, 130.43 for 8, 16, and 32 neurons), rendering them impractical due to significantly exceeding the output range of 80.27. Conversely, TDistNNs achieve markedly narrower intervals (medians: 34.69, 21.69, 18.86), demonstrating a significant improvement in prediction precision. Furthermore, TDistNNs exhibit superior interval widths compared to QuantileNNs at 16 and 32 hidden neurons, while maintaining or exceeding QuantileNN's coverage. Consequently, TDistNNs demonstrate a significantly improved trade-off between coverage and interval width on this challenging real-world data set, a direct result of its ability to effectively model heavy-tailed data.

\begin{figure}[htbp]
    \centering
    \includegraphics[width=\linewidth]{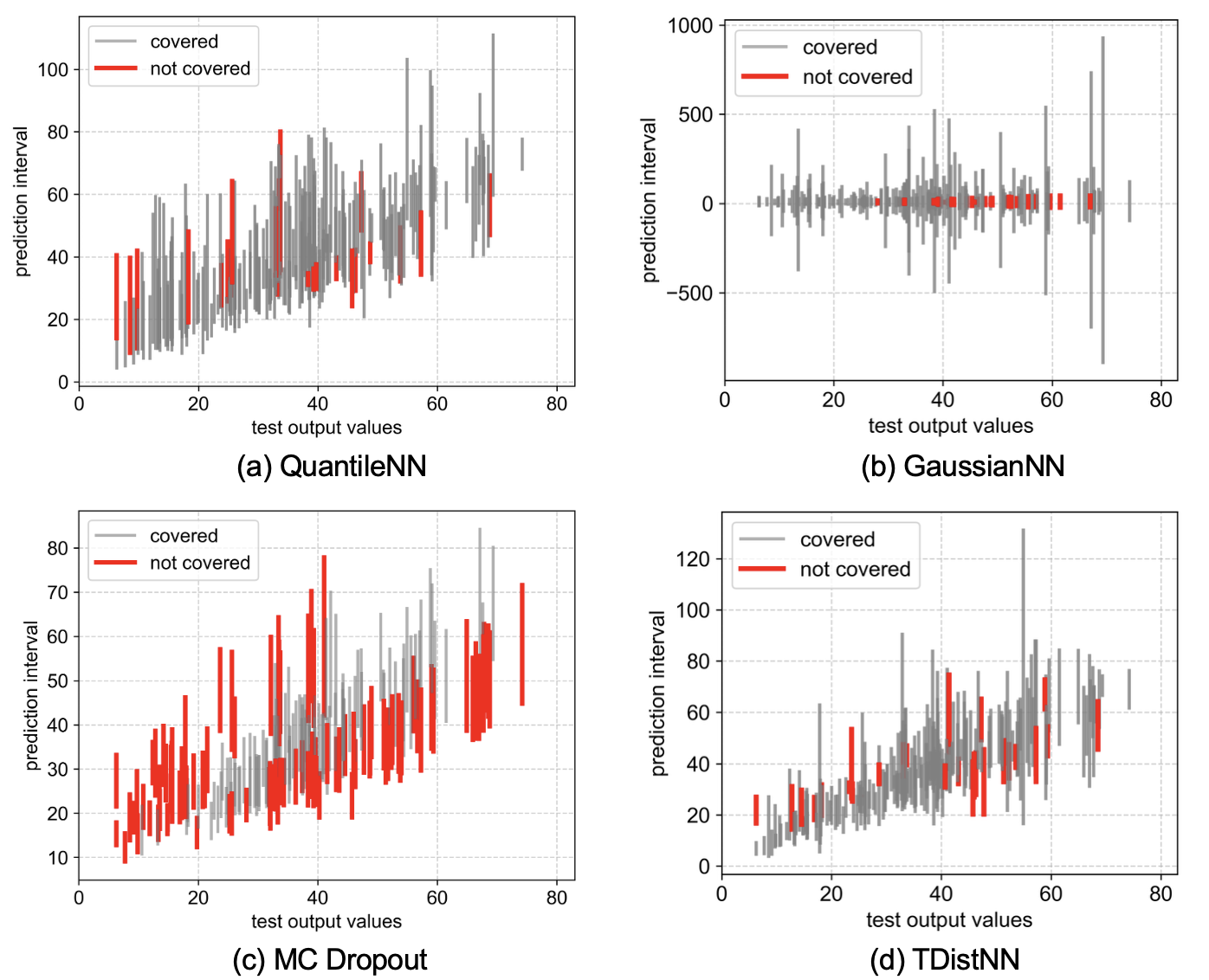}
    \caption{Graphical representation of prediction intervals for the Concrete Compressive Strength data set. The x-axis represents the true output values, and the y-axis displays the corresponding lower and upper bounds of each prediction interval.}
    \label{fig:concrete_intervals}
\end{figure}

To transition from aggregate performance analysis to a more granular examination of prediction interval behavior, we focus on a single experimental run with 16 hidden units. This approach allows for a direct visual comparison of predicted intervals with individual test set outputs, providing a closer look at the models' adaptability. The results are visualized in Fig.~\ref{fig:concrete_intervals}, where the true output value for each test point is plotted on the x-axis, and the corresponding prediction interval is represented vertically, enabling easy observation of interval bounds on the y-axis.

As anticipated from our earlier analysis using boxplots, QuantileNN generates a reasonable set of prediction intervals, with gray indicating test points where the true output falls within the interval and red highlighting instances of undercoverage. While QuantileNN's intervals generally capture the overall trend, a notable limitation arises when the true output is below 20. In these cases, the upper bounds of the prediction intervals can exceed 60, indicating a lack of fine-grained adaptability. This suggests that while QuantileNN captures general trends, it struggles to provide precise intervals for lower output values.

This issue is significantly exacerbated with GaussianNN. As Fig.~\ref{fig:concrete_intervals}(b) illustrates, some prediction intervals are entirely unreasonable, with upper bounds reaching values close to 1,000 and lower bounds dipping into negative territory\textemdash an impossibility given the strictly positive nature of the outputs. This observation aligns with our previous boxplot analysis, which demonstrated median average prediction interval widths well above 100, highlighting GaussianNN's tendency to produce excessively wide and unrealistic intervals at the cost of reaching the target coverage level. Moreover, Fig.~\ref{fig:concrete_intervals}(c) illustrates that, despite producing more meaningful prediction intervals than GaussianNN, the resulting intervals exhibit clear undercoverage, with many failing to contain the ground-truth values.

In contrast, Fig.~\ref{fig:concrete_intervals}(d) reveals that TDistNN's constructed prediction intervals exhibit a stronger correlation with the true output values. For instance, when the true output is below 20, the majority of upper bounds remain below 40. Similarly, with a single exception, the upper bounds of TDistNN's intervals consistently stay below 91.10\textemdash a marginal improvement over QuantileNN but a substantial advantage over GaussianNN. 

\begin{figure}[htbp]
    \centering
    \includegraphics[width=\linewidth]{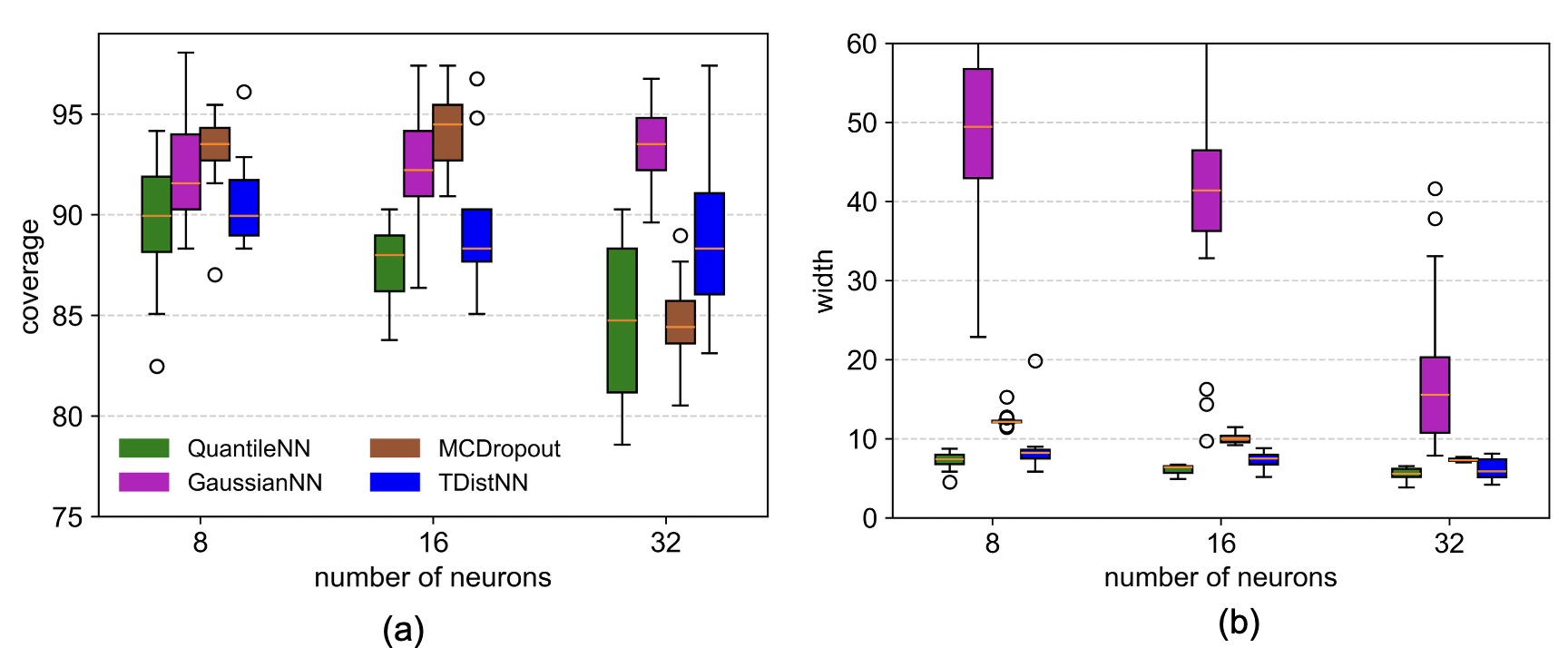}
    \caption{Boxplots illustrate (a) coverage score and (b) interval width for different architectures across 20 trials on the Energy Efficiency data set. For each network size, the results are shown in the following left-to-right order: QuantileNN, GaussianNN, MC Dropout, and TDistNN.}
    \label{fig:efficiency_cov}
\end{figure}

Next, we evaluate the performance of these methods on the Energy Efficiency data set, where the output values range from 6.01 to 43.10. Fig.~\ref{fig:efficiency_cov}(a) presents the coverage scores, while Fig.~\ref{fig:efficiency_cov}(b) illustrates the corresponding interval widths across 20 independent trials. With respect to coverage, GaussianNN attains the highest values, consistently surpassing the 90\% target across all network sizes. In contrast, MC Dropout exhibits another limitation: although its coverage is relatively high for networks with 8 and 16 neurons, it drops sharply when the network size increases to 32 neurons, indicating that its performance is sensitive to the choice of network architecture.

TDistNNs provide reasonable coverage, with median coverage scores of 89.93\%, 88.31\%, and 88.31\% for hidden layer sizes of 8, 16, and 32 neurons, respectively. Notably, TDistNN's maximum coverage values (96.10\%, 96.75\%, and 97.40\%) indicate its ability to closely approach or even surpass the target coverage level. In contrast, QuantileNNs exhibit a progressive decline in coverage as the hidden layer width increases, with median coverage scores of 89.93\%, 87.98\%, and 84.74\%, suggesting potential instability in larger network configurations.

Despite GaussianNN's superior coverage scores, a closer examination reveals that these are largely a consequence of excessively wide prediction intervals, mirroring our findings from the previous data set. As depicted in Fig.~\ref{fig:efficiency_cov}(b), GaussianNN's median interval widths are 49.45, 41.41, and 15.57 for 8, 16, and 32 neurons, respectively. While the median for 32 neurons is comparatively better, the maximum interval width of 41.63 remains problematic, as it surpasses the data set's total output range. Conversely, both TDistNN and QuantileNN generate substantially narrower and more reasonable prediction intervals. TDistNN, specifically, achieves median interval widths of 8.21, 7.51, and 5.90 for 8, 16, and 32 neurons, respectively\textemdash significantly lower than the maximum output of 43.10. This translates to TDistNN reducing the interval width by a factor of 2.64 compared to GaussianNN when using 32 neurons, all while maintaining appropriate coverage scores.

It is also worth noting that QuantileNN produces the smallest interval widths, which is generally desirable. For instance, its median interval widths are 7.42, 6.39, and 5.60, slightly lower than those achieved by our proposed TDistNN. However, as observed in Fig.~\ref{fig:efficiency_cov}(a), QuantileNN fails to provide adequate coverage scores. Therefore, we conclude that TDistNN offers the best trade-off between coverage and interval width for these fixed architectures in this case study.

We next visualize the constructed prediction intervals for individual test points using a single hidden layer with 16 units. As in the previous case, true output values are plotted along the x-axis, while the corresponding prediction intervals are represented as vertical bars, with their lower and upper bounds inferred from the y-axis.

\begin{figure}[htbp]
    \centering
    \includegraphics[width=\linewidth]{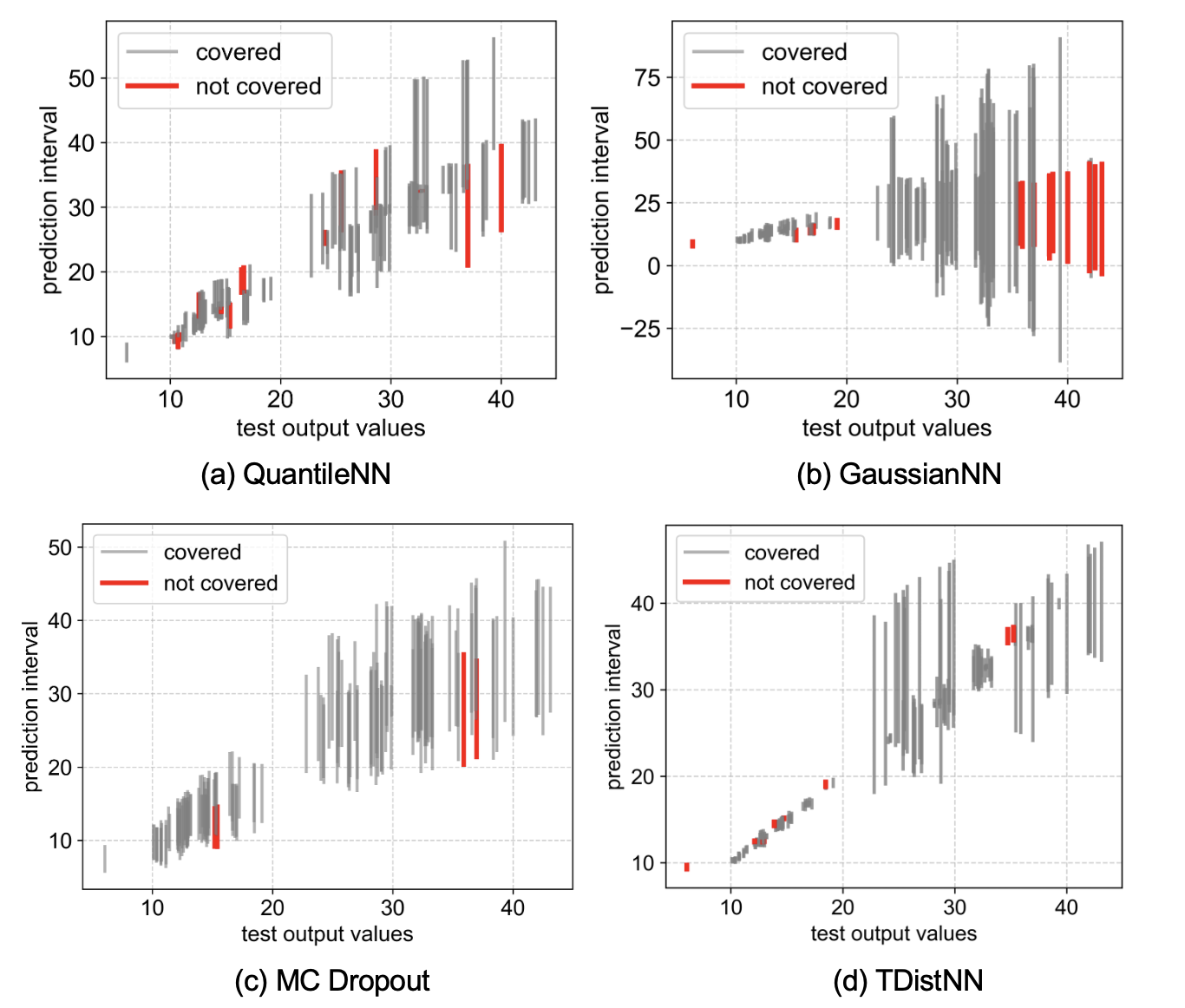}
    \caption{Graphical representation of prediction intervals for the Energy Efficiency data set.}
    \label{fig:efficiency-interval}
\end{figure}

Fig.~\ref{fig:efficiency-interval} presents a comparative visualization of prediction intervals generated by the four methods across varying true output values. Notably, all four methods demonstrate strong performance when the true output values fall below 20. In this regime, however, the proposed TDistNN consistently produces the narrowest prediction intervals. This is particularly noteworthy given that the number of instances where the true target value falls outside the predicted interval is comparable across all four methods. This indicates that TDistNN achieves greater precision without sacrificing coverage.

When the true output values exceed 20, the performance characteristics diverge. GaussianNN exhibits a tendency to generate excessively wide prediction intervals, often resulting in lower bounds that extend into the negative region. While a post-processing step could be employed to truncate these negative bounds, it is advantageous to obtain meaningful prediction intervals directly from the model, as demonstrated by TDistNN and QuantileNN. These methods inherently produce intervals that align with the positive domain of the target variable, eliminating the need for manual adjustments and improving the overall reliability of the prediction process.

\subsection{Effect of Network Depth and Cross-Validation}

In the final experiment of this section, we analyze a larger data set, the Student Performance Index data set \cite{rastiti2024}, which comprises 10,000 samples with various predictors related to academic performance. The data set includes variables such as Hours Studied, Previous Scores, Extracurricular Activities, Sleep Hours, and Sample Question Papers Practiced. The target variable, the Performance Index, serves as a measure of each student's overall academic achievement, ranging from 10 to 100, with higher values indicating better performance. 

Due to the larger size of this data set, we move beyond single-hidden-layer networks and consider deeper architectures with up to three hidden layers. We also vary the number of neurons in each hidden layer within the set 
$\{8,16,32\}$. Rather than relying on a single train-test split, we adopt a 5-fold cross-validation strategy and report the mean coverage and prediction interval width across folds for all four methods. In addition, we report the training time and prediction interval inference time as baseline measures of computational cost.

Because the loss function associated with the t-distributed output requires estimating the degrees-of-freedom parameter and computing additional elementwise operations (e.g., log, exp, softplus, and gamma functions), a modest computational overhead is expected, particularly for very small models (e.g., a single hidden layer with few neurons). By scaling up both the data set size and network capacity, this experiment provides deeper insight into the behavior of the studied methods.

\begin{table}[t]
    \centering
    \caption{Comparison of coverage, prediction interval width, and computational time on the Student Performance Index data set. Results are reported as mean values across 5-fold cross-validation. Computational time is measured in seconds. For each row, the method achieving the lowest prediction interval width is highlighted.}
    \vspace{2mm}
    \label{tab:coverage_width}
    \renewcommand{\arraystretch}{1.35}
    \setlength{\tabcolsep}{2pt}

    \begin{tabular}{c c c c c c c c c c c c c c}
        \toprule
        & & \multicolumn{3}{c}{QuantileNN}
          & \multicolumn{3}{c}{GaussianNN}
          & \multicolumn{3}{c}{MC Dropout}
          & \multicolumn{3}{c}{TDistNN} \\
        \cmidrule(lr){3-5} \cmidrule(lr){6-8} \cmidrule(lr){9-11} \cmidrule(lr){12-14}
        Dep. & Neurons
        & Cov. & Width & Time
        & Cov. & Width & Time
        & Cov. & Width & Time
        & Cov. & Width & Time \\
        \midrule

        \multirow{3}{*}{1}
        & 8  & 89.94  & 10.60  & 1.26 
             & 90.40 & 198.39  & 0.76
             & 99.74 & 29.57 & 0.77 
             & 90.00 & \textbf{6.76}  & 1.39 \\
        & 16 & 89.77 & 8.32 & 1.40
             & 91.87 & 199.03 & 0.82 
             & 99.55 & 23.86 & 1.11 
             & 89.79 & \textbf{6.75} & 1.51 \\
        & 32 & 89.62 & 7.66 & 1.38 
             & 93.90 & 185.29 & 0.92 
             & 98.63 & 17.16 & 1.57 
             & 89.80 & \textbf{6.72} & 1.57 \\
        \midrule

        \multirow{3}{*}{2}
        & 8  & 89.87 & 6.83 & 1.83 & 92.78 & 134.67 & 1.13 & 98.78 & 35.87 & 1.46 & 89.80 & \textbf{6.73} & 1.81 \\
        & 16 & 89.46 & 6.78 & 2.03 & 96.99 & 165.92 & 1.17 & 99.18 & 29.36 & 2.03 & 91.87 & \textbf{6.78} & 1.81 \\
        & 32 & 89.23 & 6.72 & 2.30 & 96.07 & 94.77 & 1.37 & 99.27 & 22.76 & 3.08 & 89.61 & \textbf{6.70} & 2.06 \\
        \midrule

        \multirow{3}{*}{3}
        & 8  & 89.97 & \textbf{6.82} & 2.62 & 97.73 & 119.79 & 1.52 & 97.39 & 36.23 & 2.25 & 95.73 & 10.97 & 2.17 \\
        & 16 & 89.28 & \textbf{6.75} & 2.90 & 93.75 & 59.26 & 1.61 & 98.64 & 31.53 & 3.04 & 91.65 & 7.44 & 2.26 \\
        & 32 & 90.01 & 7.49 & 3.51 & 89.55 & 8.98 & 2.00 & 99.47 & 25.26 & 4.84 & 92.73 & \textbf{7.02} & 2.71 \\
        \bottomrule
    \end{tabular}
\end{table}

The results for a target coverage level of 90\% are presented in Table \ref{tab:coverage_width}. We then summarize key observations about the performance of the four models across different network depths and hidden-layer widths.
\begin{itemize}
\item QuantileNNs achieve consistent coverage scores ranging from 89.23\% to 90.01\% across all network configurations. Regarding interval width, the prediction intervals produced by QuantileNNs are reasonable relative to the maximum true output value of 100. Notably, when the network depth is fixed at 1, the mean interval width decreases as the number of neurons increases, from 10.60 (with 8 neurons) to 7.66 (with 32 neurons). In contrast, we observe less noticeable differences in interval widths when the network depth is increased to 2 or 3. Importantly, increasing the depth and the number of neurons per hidden layer does not necessarily always lead to smaller prediction intervals. For example, when the depth is 3 and the number of neurons is 32, the average predicted interval width increases to 7.49. This indicates that the performance of QuantileNNs is influenced by the architecture of the underlying neural network.
\item Despite the larger sample size of this dataset, GaussianNNs continue to exhibit difficulties in achieving a suitable balance between coverage and interval width. Although the coverage scores exceed the target level of 90\%, the corresponding interval widths remain excessively large relative to the maximum true output value, indicating overly inflated prediction intervals. Interestingly, only one configuration with depth 3 and 32 neurons yields an average predicted interval width below 10, specifically 8.98. While this width is still larger than those produced by other models, such as QuantileNNs, it suggests that increasing the network depth or capacity can help mitigate the overestimation of variance in GaussianNNs. However, the downside is that the performance of GaussianNNs is overly sensitive to the network architecture. 
\item Unlike previous experiments in which MC Dropout suffered from undercoverage, we observe a substantial increase in the mean coverage score across all configurations in this setting, with the minimum coverage reaching 97.39\%, well above the target level of 90\%. This improvement in coverage comes at the cost of significantly wider prediction intervals, which are approximately two to three times larger than those produced by QuantileNNs. One notable trend is that, for a fixed network depth, increasing the number of neurons tends to result in smaller prediction intervals, which is more desirable. Despite this trend, the minimum observed prediction interval width is 17.16, which remains uncompetitive compared to the other models under study. This also indicates that the performance of MC Dropout depends strongly on the network configuration and requires careful tuning.
\item The proposed TDistNN consistently achieves coverage scores close to or exceeding the target coverage level of 90\% across all tested configurations. In contrast to the other models, the mean prediction interval widths are considerably more stable, remaining in the range of 6 to 7 for most configurations, with only one case reaching a maximum value of 10.97. Furthermore, for each configuration, corresponding to each row of Table \ref{tab:coverage_width}, we highlight the model that achieves the smallest mean prediction interval width. In this comparison, TDistNN yields the smallest interval width in 7 out of 9 cases. Overall, these results indicate that the performance of TDistNN is less dependent on architectural configurations, which is a desirable property for practical applications where extensive hyperparameter tuning may be impractical. Across all configurations, TDistNN also achieves the minimum observed interval width of 6.70.
\item We also report the total computational time required to train each model across five runs corresponding to 5-fold cross-validation, together with the inference time. As expected, for all models, increasing the network size, either by increasing the depth or the number of neurons per hidden layer, leads to higher computational cost. When comparing different methods, GaussianNN is the fastest model, which can be attributed to its relatively simple loss function, the availability of a built-in loss implementation in PyTorch, and the fact that it does not require repeated training or sampling during inference, unlike QuantileNNs and MC Dropout.
In terms of scalability as the network size increases, MC Dropout is the most computationally expensive approach due to the need to generate multiple stochastic forward passes at inference time. In our experiments, we use $T=100$ realizations, although this value can be adjusted to trade off accuracy and computational cost. While TDistNN is slightly more time-consuming than GaussianNN for a fixed configuration, a key advantage of TDistNN is that it achieves competitive coverage and prediction interval widths using smaller networks. For example, TDistNN requires 1.57 seconds to produce a mean interval width of 6.72 using a network with depth 1 and 32 neurons, whereas GaussianNN requires approximately 2 seconds to produce a wider prediction interval of 8.98 using a deeper network with depth 3 and 32 neurons.
\end{itemize}

Consequently, TDistNNs offer a robust and effective solution for practical uncertainty quantification, delivering accurate coverage and consistently narrow prediction intervals while maintaining stable performance across different network architectures.

\section{Conclusion}
\label{sec:conclusion}
While probabilistic neural networks (PNNs) offer a valuable approach to quantifying uncertainty in model predictions by parameterizing predictive distributions, their reliance on restrictive assumptions, such as Gaussianity, poses a significant limitation, particularly in real-world regression problems prone to outliers and extreme values. This paper demonstrated the efficacy of employing a more flexible predictive distribution, specifically the t-distribution parameterized by location, scale, and shape/degrees of freedom. The proposed TDistNN framework provides enhanced control over tail behavior, resulting in prediction intervals that strike a superior balance between coverage and width, crucial for reliable and informed decision-making. We detailed the TDistNN architecture, the derivation of a tailored loss function, and the training process for generating prediction intervals, showcasing its practical applicability.

This work, by establishing a principled framework for flexible PNNs, opens several promising avenues for future research. First, while we maintained fixed architectures in this study for fair comparison, future work will explore specialized hyperparameter tuning techniques designed for TDistNNs. These techniques will optimize the trade-off between coverage and prediction interval width by incorporating metrics that consider the parameterized distribution. Second, we will investigate advanced regularization strategies, such as imposing additional constraints on the t-distribution parameters to control the complexity of the degrees of freedom. Third, we aim to extend the TDistNN framework to selective regression, enabling the model to reject predictions when confidence is low. This will involve optimizing the trade-off between prediction interval width, coverage score, and rejection rate, a critical aspect for deploying regression models in real-world applications.
%\newpage
%\setlinespacing{1}
\bibliographystyle{plain}
%\bibliographystyle{ieeetr}
%\bibliographystyle{unsrt}
%\bibliography{krr}
\bibliography{refs}

\end{document}